# Awal – Community-Powered Language Technology for Tamazight

Alp Öktem[1,2], Farida Boudichat[2]

[1]Col·lectivaT
alp@collectivat.cat

[2]Awal Team
awal@collectivat.cat

## Abstract.

This paper presents Awal (ⴰⵡⴰⵍ), a community-powered initiative for developing language technology resources for Tamazight. We provide a comprehensive review of the NLP landscape for Tamazight, examining recent progress in computational resources, and the emergence of community-driven approaches to address persistent data scarcity. Launched in 2024, awaldigital.org platform addresses the underrepresentation of Tamazight in digital spaces through a collaborative platform enabling speakers to contribute translation and voice data. We analyze 18 months of community engagement, revealing significant barriers to participation including limited confidence in written Tamazight and ongoing standardization challenges. Despite widespread positive reception, actual data contribution remained concentrated among linguists and activists. The modest scale of community contributions —6,421 translation pairs and 3 hours of speech data—highlights the limitations of applying standard crowdsourcing approaches to languages with complex sociolinguistic contexts. We are working on improved open-source MT models using the collected data.

## ⵜⴰⴳⴹⵡⵉⵜ

ⴰⵔⵣⵣⵓ ⴰⴷ ⴰⵔ ⵉⵙⵎⵉⵍⵉ ⴰⵡⴰⵍ, ⵢⴰⵏ ⵓⵙⵏⵜⵉ ⵉⵖⵍⵉⵏ ⵙ ⵜⵣⵎⵔⵜ ⵏ ⵜⵎⵓⵏⵜ ⵉ ⵓⵙⵅⵓⵖⵍⵓ ⵏ ⵉⵙⵓⴳⴰⵎⵏ ⵏ ⵉⵜⵉⴽⵏⵓⵍⵓⵊⵉⵏ ⵏ ⵜⵓⵜⵍⴰⵢⵜ ⵉ ⵜⵎⴰⵣⵉⵖⵜ. ⴳ ⴰⵔⵣⵣⵓ ⴰⴷ, ⵔⴰ ⵏⵙⵔⵙ ⵢⴰⵏ ⵓⵣⵣⵔⴰⵢ ⴰⴳⴰⵜⴰⵔ ⵏ ⵓⴷⵖⴰⵔ ⵏ ⵓⵙⵎⴽⵍ ⵏ ⵜⵓⵜⵍⴰⵢⵜ ⵜⴰⴳⴰⵎⴰⵏⵜ (NLP) ⵉ ⵜⵎⴰⵣⵉⵖⵜ, ⴷ ⵔⴰ ⵏⵖⵙ ⵜⴰⵍⴰⵢⵉⵏ ⵜⵉⵎⴰⵢⵏⵓⵜⵉⵏ ⴳ ⵢⵉⴳⵔ ⵏ ⵉⵙⵓⴳⴰⵎⵏ ⵏ ⵉⵎⵙⵙⵓⴷⵙⴰⵏⵏ (computational resources), ⴷ ⵏ ⵓⵙⵏⵀⵍ ⵏ ⵜⵖⴰⵔⴰⵙⵉⵏ ⵉⵅⴷⴷⴰⵏ

ⵖⴼ ⵜⴳⵔⴰⵡⵜ ⵢⴰⴽ ⴰⴷ ⵏⵙⵎⵍ ⵜⴰⴷⵔⵙⵉ ⵏ ⵓⵎⵍⴰⵏ ⵉⵜⵜⴰⴷⵔⴰⵏ. ⵉⵜⵜⵓⵙⵏⵎⴰⵍⴰ ⵓⵙⵏⵜⵉ ⵏ awaldigital.org ⴳ ⵓⵙⴳⴳⵯⴰⵙ ⵏ 2024. ⵉⵙⵔⵙ ⵓⵙⵏⵜⵉ ⴰⴷ ⵜⵉⵃⵃ ⵏⵏⵙ ⵖⴼ ⵜⴷⵔⵙⵉ ⵏ ⵜⵉⵍⵉⵜ ⵏ ⵜⵢⵜⵍⴰⵢⵜ ⵜⵎⴰⵣⵉⵖⵜ ⴳ ⵓⵎⴰⴹⴰⵍ ⴰⵎⵓⵃⵃⵓⵏ. ⵏⴽⵙⵔ ⵢⴰⵜ ⵜⴷⴰⴱⵓⵜ ⵜⴰⵎⴰⵡⴰⵙⵜ ⴷⴰ ⵢⵓⵊⵊⴰⵏ ⵉⵎⵙⴰⵡⴰⵍⵏ ⵏ ⵜⵎⴰⵣⵉⵖⵜ ⴰⴷ ⵙⵙⴽⵛⵎⵏ ⴰⵙⵓⵖⵍⵏ ⴷ ⵉⵎⵙⵍⵉ ⵏⵏⵙ. ⴰⵔ ⵏⵙⵙⴼⵙⴰⵢ 18 ⵏ ⵡⴰⵢⵢⵓⵔⵏ ⵏ ⵜⴷⵔⴰⵡⵜ ⵜⴰⵎⴳⵔⴰⵡⵜ. ⵜⴰⴷⵔⴰⵡⵜ ⴰⴷ ⵜⴰⵡⵙ ⴰⵖ ⴰⴷⴷ ⵏⵙⵙⵓⴼⵖ ⵉⵔⴼⵜⵉⵍⵏ ⵉⵅⴰⵜⴰⵔⵏ ⴷⴰ ⵉⵜⵜⴱⴷⴷⴰⵏ ⵎⵍⵉⴷ ⵜⴳⵔⴰⵡⵜ. ⵙⴳ ⵉⵔⴼⵜⵉⵍⵏ ⴰⴷ ⵜⴷⵔⵙⵉ ⵏ ⵜⵉⵔⵔⴰ ⵙ ⵜⵎⴰⵣⵉⵖⵜ ⴷ ⵜⵣⵎⵎⴰⵔ ⵏ ⵓⵙⵏⴰⵡⴰⵢⴰ ⵏ ⵜⵓⵜⵍⴰⵢⵜ. ⵡⴰⵅⵅⴰ ⴰⵙⵏⵜⵉ ⴰⴷ ⵉⵏⵏⵓⵍ ⵢⴰⵏ ⵓⵙⵏⵓⴱⴳ ⵓⵎⵏⵉⴳⵏ, ⵜⴰⴷⵔⴰⵡⵜ ⵜⴰⵎⴳⵔⴰⵡⵜ ⵡⴰⵙⴰ ⵜⵍⵍⴰ ⵜⴷⵔⵓⵙ ⴰⵀⴰ ⵜⴷⴷⴰ ⴷ ⵖⴰⵙ ⵙⴳ ⴷⴰⵔ ⵉⵎⵙⵏⵉⵍⵙⵏ ⴷ ⵉⵎⵖⵏⴰⵙⵏ. ⵡⴰⵅⵅⴰ ⵜⴷⵔⵓⵙ, ⵜⴰⴷⵔⴰⵡⵜ ⵜⴰⵎⴳⵔⴰⵡⵜ - ⴷⴰ ⵉⵙⵎⵓⵏⵏ ⴽⵔⴰ ⵏ 6,421 ⵏ ⵉⴽⵏⵉⵡⵏ ⵏ ⵓⵙⵓⵖⵍ ⴷ 3 ⵜⵙⵔⴰⴳⵉⵏ ⵏ ⵉⵎⵙⵍⵉ - ⵜⵙⵎⵍⴰ ⴷ ⵉⵔⴼⵜⵉⵍⵏ ⴷⴰ ⵉⴱⴷⴷⴰⵏ ⵎⵉⵍⴷ ⵏ ⵓⵙⵔⵙ ⵏ ⵜⵉⵎⵢⴰⴷⴰⵙⵉⵏ ⵜⵉⵏⴰⵡⴰⵢⵏ ⵖⴼ ⵜⵖⵔⵉ ⵏ ⵜⵡⵉⵙⵉ ⵏ ⵜⵓⵜⵍⴰⵢⵉⵏ ⴷⴰⵔ ⵎⵉ ⵍⵍⴰⵏⵜ ⵉⵎⵏⴰⴹⵏ ⵉⵙⵏⵎⵓⵏⵉⵍⵙⵜⵏ ⵎⵢⴰⵏⴰⵡⵏ. ⵖⴼ ⴰⴽⴽⵯ ⵓⵢⴰ, ⴰⵔ ⵏⵙⵡⵓⵔⵉ ⵖⴼ ⵓⵙⴼⵓⵍⴽⵉ ⵏ ⵉⵣⵓⵔⴰⵜ ⵏ ⵜⵙⵓⵖⵍⵜ ⵜⵉⵎⵉⵙⵜ ⵉⵜⵜⵢⴰⵏⴰⴼⵏ ⵉ ⵓⴳⴷⵓⴷ ⵙ ⵜⵎⴰⵡⴰⵙⵜ ⵏ ⵓⵎⵍⴰⵏ ⵉⵜⵜⴰⵙⵎⵓⵏⵏ.

## 1. Introduction

The digital representation of the world's languages remains deeply unequal, with technological advances primarily benefiting a small subset of widely-spoken languages while thousands of others remain marginalized in digital spaces. This digital language divide is particularly pronounced for African languages, where despite representing over 2,000 languages spoken by more than a billion people, computational resources and language technologies remain scarce (Nekoto et al., 2020; Kreutzer et al., 2022). Among these underrepresented languages, Tamazight (also known as Amazigh or Berber) stands as a compelling case study of both the challenges and possibilities of community-driven language technology development.

Tamazight, spoken by over 40 million people across North Africa and diaspora communities worldwide, has experienced a remarkable journey from marginalization to official recognition in multiple countries since 2011. However, this political recognition has not automatically translated into robust digital presence or computational resources. Despite recent progress—including the launch of Tamazight Wikipedia in 2023, the inclusion of Tamazight in Google Translate, and the availability of digital tools by IRCAM and other linguists and developers—the language still lacks sufficient presence on the web to power AI-based technologies, especially large language models (LLMs). While current initiatives focus on standardization and promoting writing skills, parallel efforts are needed to create high-quality digital content and structured data that can train effective AI systems.

This paper presents the Awal (ⴰⵡⴰⵍ) project, a community-driven initiative launched in 2024 by three organizations in Catalonia to foster the development of language technologies for Tamazight through crowdsourced data collection. Awal collects translation pairs and transcribed speech data directly usable for training AI models such as automatic speech recognition (ASR), machine translation (MT), and large language models (LLMs). The platform encourages community participation through gamification elements, while maintaining quality through peer validation mechanisms.

The paper is structured as follows. In Section 2, we provide a comprehensive review of the NLP landscape for Tamazight, examining its linguistic profile and sociopolitical status (Section 2.1) and the challenges and progress in Tamazight NLP (Section 2.2), and community-driven approaches (Section 2.3). Section 3 details the Awal project, covering its inception and objectives (Section 3.1), the awaldigital.org platform (Section 3.2), voice data collection through Common Voice (Section 3.3), community engagement and outreach campaigns (Section 3.4), current statistics and data access (Section 3.5), and limitations, challenges and way forward (Section 3.6). We finally conclude in Section 4.

## 2. Landscape in Tamazight NLP

This section examines the current state of Tamazight in the digital realm. We first provide an overview of the language's linguistic diversity and sociopolitical evolution, then review existing computational work and resources, before discussing the emergence of community-driven approaches.

### 2.1. *Linguistic profile and sociopolitical status of Tamazight*

Tamazight, also known as Amazigh or Berber, belongs to the Afro-Asiatic language branch, with over 40 million people across a vast region in North Africa, principally in Morocco, Algeria, Tunisia, Libya, and by communities in Spain, Mauritania, Niger, Mali, and Egypt, and also within the diaspora. Figure 1 shows geographic distribution of Tamazight dialects across North Africa. Tamazight is an official language in Morocco, since 2011, Algeria, since 2016, in certain districts of Libya, since 2017, and Mali, since 2023. The language exhibits significant diversity, with notable dialects such as Tachelhit (shi), spoken in the southwest and the High Atlas, Tarifit (rif), spoken in the Rif region of Morocco, Central Atlas Tamazight (tzm), spoken in central and southeast Morocco, Kabyle (kab) spoken in Algeria, Shawiya or Tacawit (shy) also in Algeria, Tuareg Tamahaq (thv) in the southern regions of Algeria, Niger, and Mali, and Tamasheq (taq) in Mali and Niger (Lafkioui, 2018).

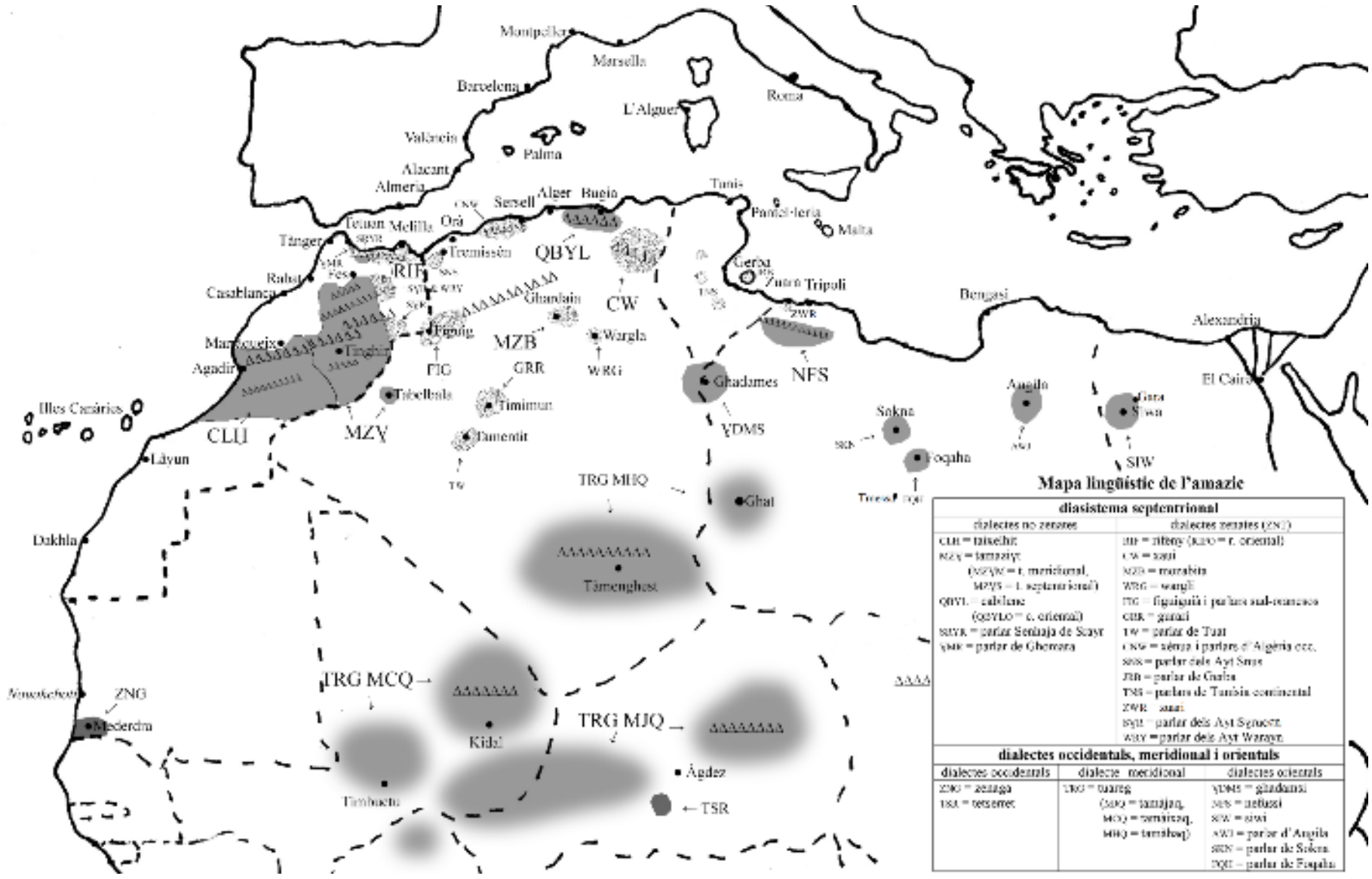

Figure 1: Geographic distribution of Tamazight dialects across North Africa (Múrcia, 2017)

Tamazight has faced marginalization under the effects of colonization and Arabization, leading to a decline in the language's status and use. Amazigh activism, particularly during the Amazigh Spring, played a crucial role in highlighting these issues and advocating for linguistic and cultural rights (Roque, 2009; CIEMEN and Casa Amaziga de Catalunya, 2019). This activism led to significant progress in Morocco, starting with the establishment of the Royal Institute of Amazigh Culture (IRCAM) in 2001, which marked the beginning of the institutionalization of Tamazight. The standardized form Standard Moroccan Tamazight (ISO-639 code: zgh) was developed by IRCAM, combining features of the three main dialects shi, tzm and rif (Boukous, 2014), as well as other variants such as Touareg. As part of the standardization efforts, the Neo-Tifinagh script, developed by IRCAM and based on the traditional Tifinagh script, was chosen as the official script for Tamazight in 2003. This modern graphical system, known as *Tifinaghe-IRCAM*, was preferred over the historically used Latin and Arabic scripts, providing a standardized and phonologically accurate writing system for the diverse Moroccan Tamazight varieties (Soulaimani, 2016; Ataa-Allah and Boulaknadel, 2012).

## 2.2. *Challenges and Progress in Tamazight NLP*

Tamazight's complexity in NLP arises from its script, morphology, and high dialectal variations. The existence of various alphabets, including Tifinagh, Latin, and Arabic, complicates standardization and computational processing. Additionally, the language's rich morphology, involving both inflectional and derivational processes, along with significant dialectal differences, poses challenges for developing consistent NLP applications. These factors make tasks such as part-of-speech tagging, syntactic parsing, and machine translation particularly difficult (Ataa-Allah and Boulaknadel, 2012).

Despite its relatively recent standardization and entry into the digital realm, exemplified by the launch of the Tamazight Wikipedia in 2023, Tamazight NLP has shown considerable progress in recent years. Notable work includes the construction of a Standard Tamazight Corpus (Boulaknadel and Ataa Allah, 2013), the development of a morphosyntactically annotated corpus (Amri et al., 2017), and advancements in Amazigh word embedding (Faouzi et al., 2023). Additionally, a tool for Tamazight verb conjugation has been developed by Ataa-Allah and Boulaknadel (2014), concordancer and Tifinagh-adapted search engine by Ataa-Allah and Boulaknadel (2012). Some of these tools and resources are made available through IRCAM's portal[1].

The most notable open-source MT work includes the NLLB project, which incorporates Tamazight among the 200 languages it supports (NLLB Team et al., 2022). This project relies on the FLORES training and evaluation sets, which, as the Awal project also intends to address, require corrections for improved accuracy (Goyal et al., 2022). SIB-200, an extension of FLORES, offers a new benchmark for topic classification across 205 languages, including Tamazight (Adelani et al., 2024). The MADLAD project, another significant development, offers a large multilingual and document-level audited dataset based on CommonCrawl, further expanding the resources available for low-resource languages like Tamazight (Kudugunta et al., 2023). Additionally, the GlotCC project provides a general domain monolingual dataset derived from CommonCrawl (Kargaran et al., 2024), covering more

1 https://tal.ircam.ma/talam/

than 1000 languages, and introduces GlotLID, an open-source language identification model supporting over 2000 labels, both of which include Moroccan Standard Tamazight as well as other Tamazight varieties (Kargaran et al., 2023).

### 2.3. *Community-Driven Approaches*

The fundamental challenge facing Tamazight and other underrepresented languages in the digital age is clear: no data means no AI. Large language models and other AI technologies depend on massive amounts of data to function effectively. For underrepresented languages like Tamazight, the solution requires both organic content creation—blogs, videos, and online resources—and direct data collection efforts specifically designed for AI training.

Community-driven initiatives have emerged as a response to the underrepresentation of languages in digital spaces. Since these languages are less present online, they rarely appear in large-scale web crawls that form the basis of modern language technologies. Community participation in data collection, similar to Wikipedia's collaborative model, has shown success in creating substantial linguistic resources. Notable examples include Common Voice (Ardila et al., 2020), Tatoeba[2], NaijaVoices for Igbo, Hausa, and Yoruba (Emezue et al., 2025) and Darija data collection[3].

These types of initiatives have demonstrated that even marginalized languages can achieve representation in both commercial language technology services and open-source models when communities engage seriously in data collection efforts (Emezue et al., 2025; Gonzalez-Agirre et al., 2024). In contrast to this participatory approach, the development of language technologies for underrepresented languages typically follows a top-down approach, driven by academic institutions or technology companies with limited community input (Moshagen et al., 2024; Bird, 2020; Schwartz, 2022). This approach risks significant harm to languages and communities by misrepresenting languages through technologies built without meaningful participation from speaker communities. As Bird (2020) warns, by treating Indigenous knowledge as a commodity, speech and language technologists risk "disenfranchising local knowledge authorities, reenacting the causes of language endangerment." The quality of training data becomes particularly critical for marginalized languages as well, where inaccurate content can

2 http://tatoeba.org

3 https://www.atlasia.ma/

distort digital representations and propagate errors through AI systems (Kreutzer et al., 2022; Lau et al., 2025).

For Tamazight specifically, several community-driven initiatives have emerged to fill the gap. The inclusion of Tamazight in Google Translate marked a significant milestone, though initial translation quality was poor. However, community feedback and participation gradually improved the system's performance, demonstrating the importance of speaker involvement in refining language technologies.

The Tamazight NLP community in Hugging Face[4] represents another grassroots effort to advance language technologies. This community of 40 members aims to develop models and tools for Tamazight in both NLP and speech technologies. In the Wikipedia ecosystem, Standard Moroccan Tamazight was launched in 2024[5], joining existing Tamazight varieties including Kabyle (launched in 2007) and Tachelhit/Shilha (launched in 2021), with other Tamazight varieties in incubation phases.

Understanding these dynamics informed the design of the Awal project, which seeks to build upon the lessons learned from existing community efforts while addressing some of their limitations.

## 3. Awal Project

The Awal project (ⴰⵡⴰⵍ, “Speech” or “Discourse” in Tamazight) represents a community initiative focused on preserving and promoting Tamazight in the digital realm. Launched in 2024 through collaboration among Catalan non-profit entities promoting Tamazight and open language technology[6], the project aims to strengthen language technologies in Tamazight with a focus on crowdsourced language data collection from native speakers.

This chapter details the Awal project's methodology, platform architecture, current achievements and its limitations. We examine both the translation data collection system implemented at awaldigital.org and the voice data collection efforts through Common Voice, analyzing their effectiveness in building language resources for Tamazight NLP development.

### 3.1. *Project Inception and Objectives*

4 https://huggingface.co/Tamazight-NLP
5 https://zgh.wikipedia.org/
6 CIEMEN, Casa Amaziga de Catalunya, and Col·lectivaT

The Awal project emerged from Catalunya, home to over 100,000 Tamazight speakers within a territory that has a history of promoting its marginalized languages Catalan and Aranese Occitan. This context comes with institutional support and community infrastructure necessary for language technology development initiatives as can be seen in initiatives Aina[7] and Araina[8].

Awal's core objectives include:

1. preserving and promoting Tamazight through innovative digital tools,
2. motivating community participation in data creation,
3. developing open-source assistive technologies including machine translation and speech recognition systems adapted to Tamazight.
4. providing educational resources for language learning, and
5. strengthening communication among speakers to reinforce cultural identity

Initial efforts focused on manual collection of translated phrases, but this approach proved insufficient for scaling to the thousands of sentence pairs needed for effective machine translation. The project therefore adopted a community-driven platform approach that encourages broader participation while maintaining quality through peer validation mechanisms.

## 3.2. *awaldigital.org Platform*

The awaldigital.org platform[9] serves as the central hub for the project. Any user can access information about the project and use the integrated machine translation application. Figure 2 shows the “Contribute” page where registered users can contribute translations where the source sentence appears on the left and the target on the right. Users select the language for both sides, with at least one required to be Tamazight, which can be written in either Tifinagh or Latin script.

7 https://projecteaina.cat/

8 https://www.projecte-araina.org/

9 https://awaldigital.org/

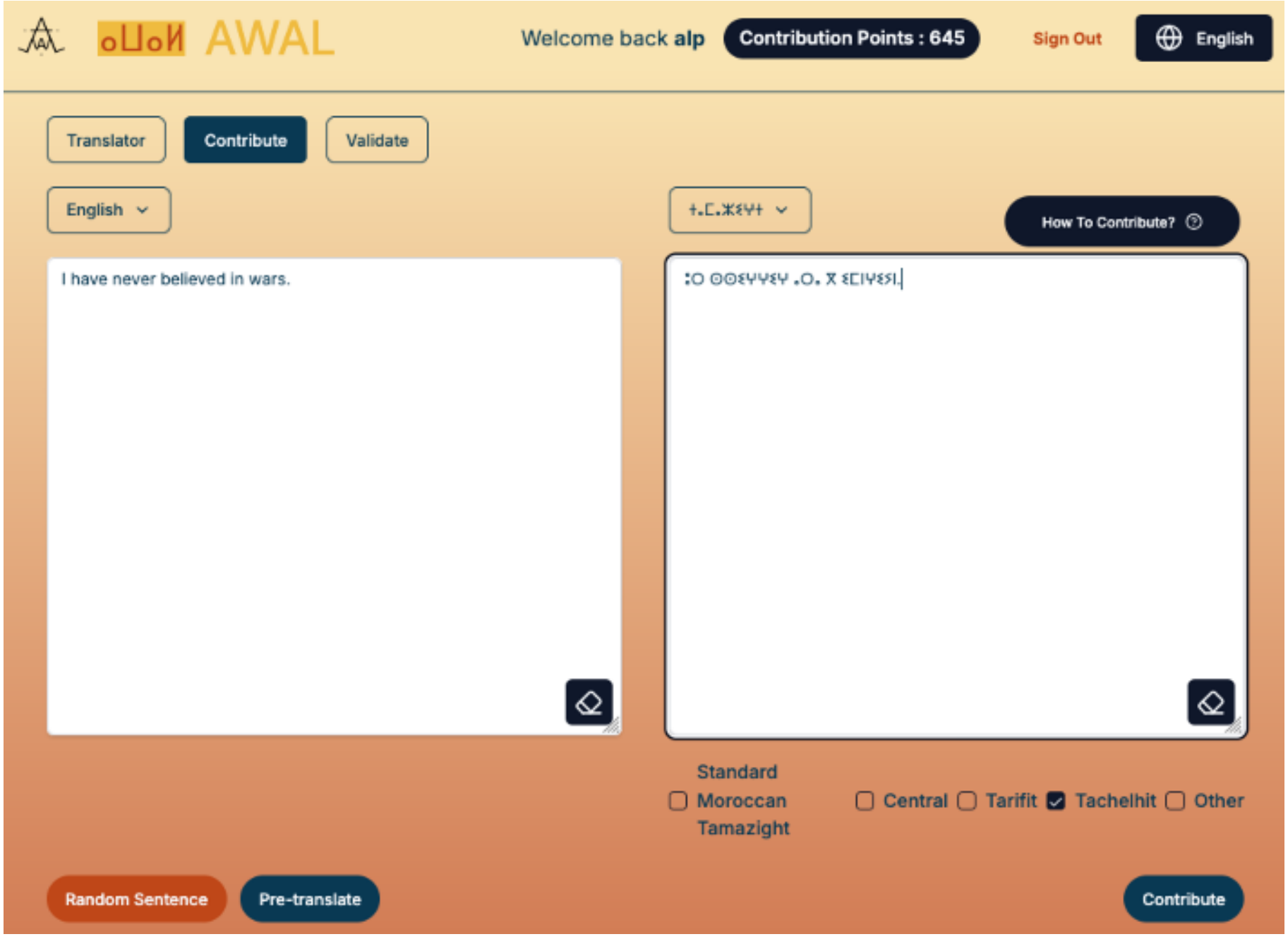

Figure 2: Contribute page allows registered users to add translations

Users can optionally mark the dialect they're writing in. The Random Sentence feature loads sentences from a database of Creative Commons-licensed texts into the source box.

The Pre-translate option automatically translates source text to the target language using the integrated machine translation model. Users must then correct this translation before submission, creating a post-editing workflow that improves efficiency.

A gamification system awards points for each character input in both source and target boxes, with users able to view their ranking on a leaderboard. This mechanism creates a friendly competition atmosphere and encourages sustained participation.

The translation interface supports bidirectional translation between Tamazight and multiple languages including Catalan, Spanish, French, Moroccan Arabic, and English. Contribution of sentence pairs are also allowed in these languages.

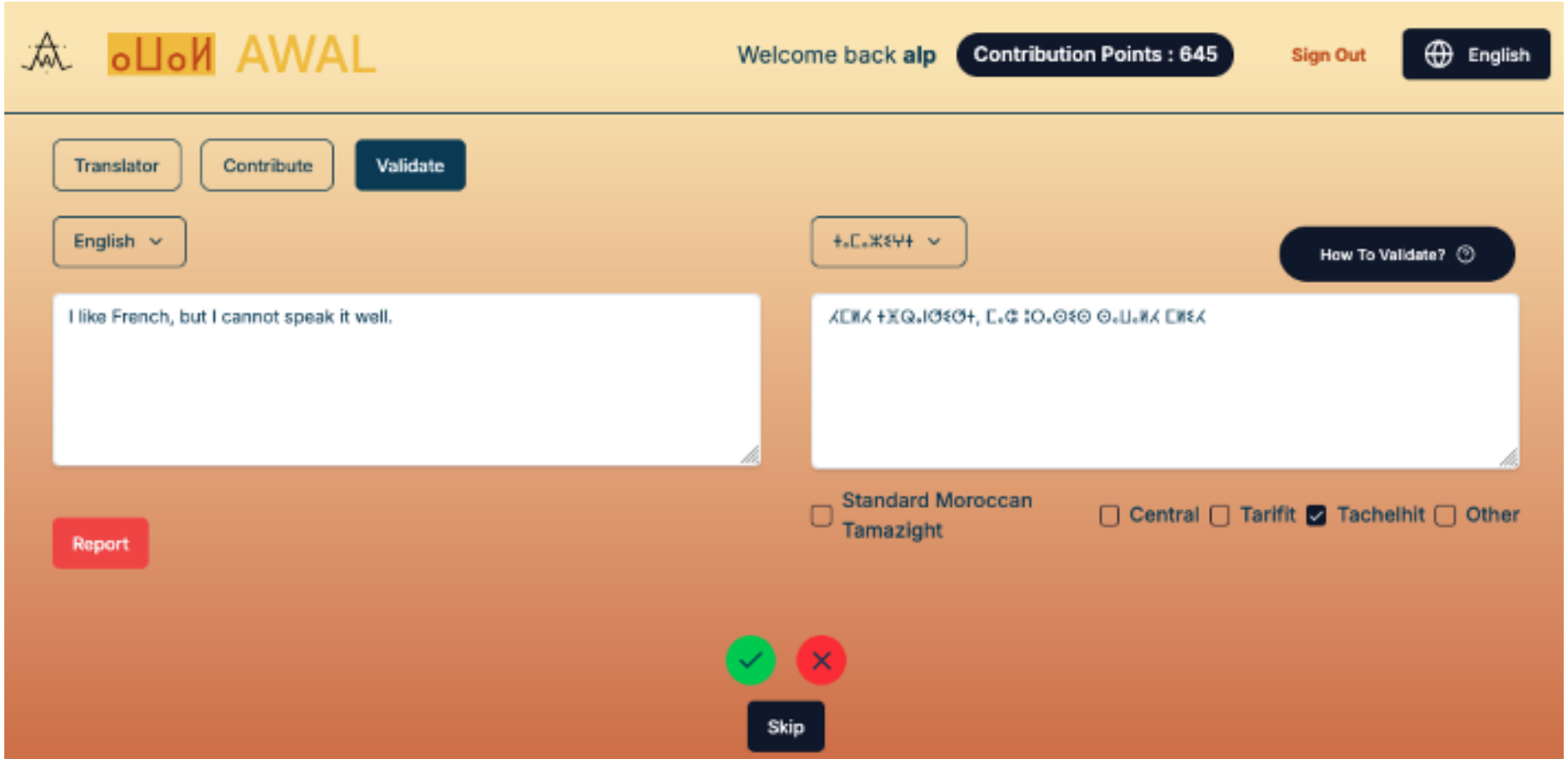

Figure 3: Validation screen.

Quality control is maintained through peer-validation (Figure 3) where users review translations submitted by others. Validators use acceptability guidelines focusing on meaning, fluency, and grammatical accuracy, with two validation approvals moving entries into the validated corpus through peer review. Two validation approvals move entries into the validated corpus.

The visual identity of the platform reflects Tamazight cultural identity and responds to the language's specific requirements, particularly its dialectal diversity and script variations. Rather than imposing strict dialectal standardization, Awal welcomes contributions from speakers of all Tamazight variants[10].

### 3.3. *Voice Data Collection through Common Voice*

Voice data collection is realized through Mozilla's Common Voice platform[11]. This integration required translating the platform interface into Tamazight and populating it with an appropriate sentence collection for recording. Figure 4 shows the recording screen. Contributors read displayed sentences aloud, creating a speech corpus for automatic speech recognition

10 The platform categorizes contributions into five variants: Standard Moroccan Tamazight, Central, Tarifit, Tachelhit, and Other. This classification represents a pragmatic compromise between differing community perspectives on dialectal standardization and necessarily excludes some varieties, particularly those outside Morocco.

11 https://commonvoice.mozilla.org/zgh

development. A similar peer-validation system also exists in Common Voice.

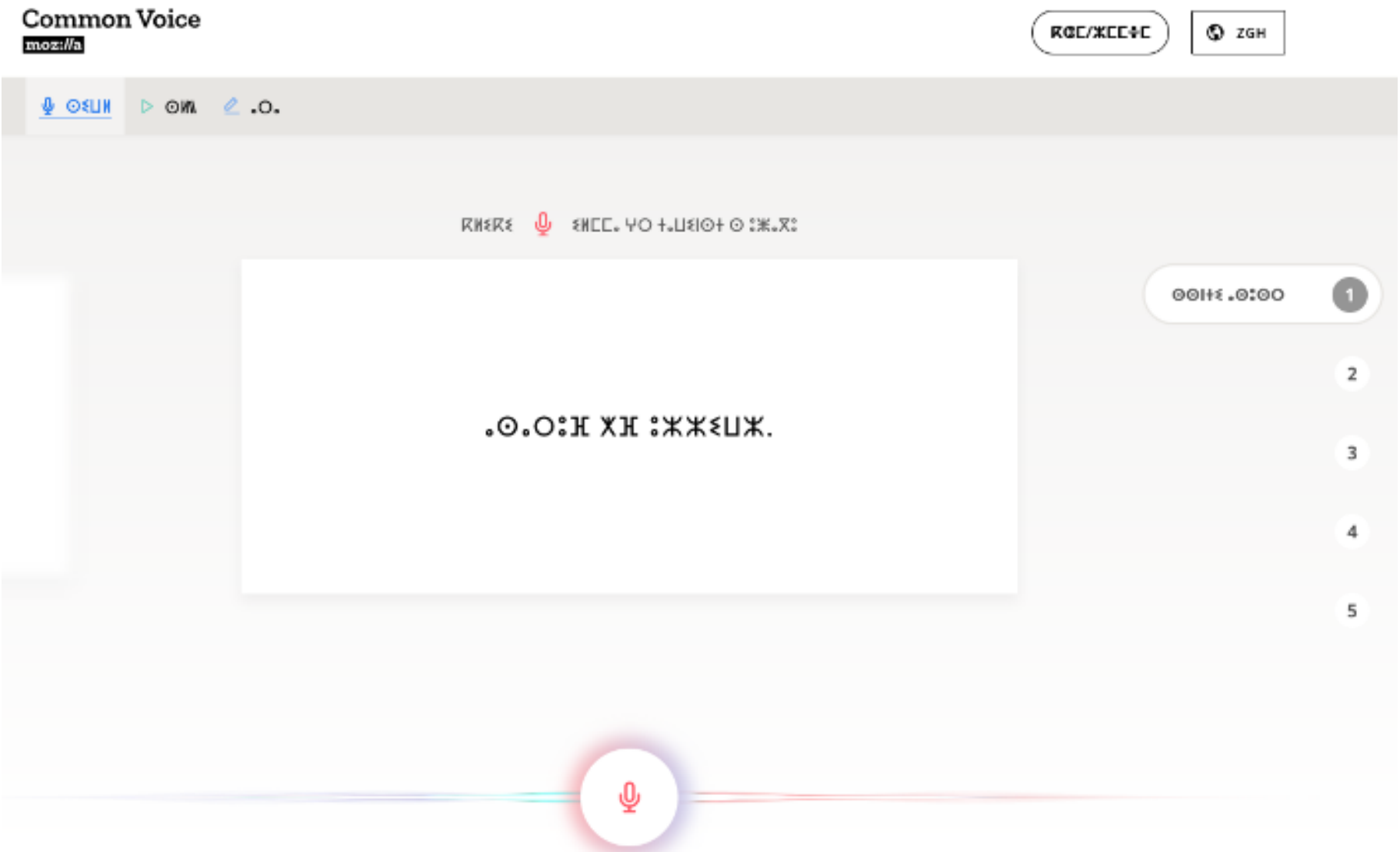

Figure 4: Tamazight voice data contribution in Common Voice.

### 3.4. *Community Engagement and Outreach Campaigns*

Our outreach involved social media campaigns across Instagram, Facebook, LinkedIn, and Telegram channels, alongside presentations at cultural events and a flagship datathon.

The Awal Datathon, held on February 17, 2024 in Barcelona, marked the project's launch event. Organized both virtually and in-person, the weekend-long datathon successfully collected over 3,500 translated sentences and 1 hour of voice recordings in Tamazight with participation from Catalonia, Morocco and different parts of Spain. Beyond this event, the project conducted targeted workshops and presentations in Bilbao, Tortosa, and Vic.

### 3.5. *Current statistics and data access*

As of 13th of June, 2025, the platform has 286 registered users, though only 66 users (23%) have actually contributed translations to the platform. The remaining users registered but did not submit any content. Since its inception in January 2024, the platform has collected 6,421 total contributions, of which 2,182 (34%) are validated. Among contributing users, the average is 86 contributions per user.

The number of translations collected for each pair is detailed in Table 1. Catalan-Tamazight (Latin) emerges as the most productive pair with 1,718 contributions, followed by English-Tamazight (Tifinagh) with 1,539 contributions and Spanish-Tamazight (Latin) with 1,203 contributions.

In addition to text translations, Awal has collected 3 hours of speech data through Common Voice platform with 2 of them validated[12].

| | Tifinagh | Latin | TOTAL |
|---|---|---|---|
| ar | 11 | 10 | 21 |
| ary | 311 | 136 | 447 |
| ca | 395 | 1718 | 2113 |
| es | 90 | 1203 | 1293 |
| en | 1539 | 693 | 2232 |
| fr | 83 | 232 | 315 |
| TOTAL | 2429 | 3992 | 6421 |

Table 1: Contribution in each language pair as of 13.06.2025.

We share all parallel and monolingual text data through our Hugging Face repository[13] while the voice data can be directly accessed from Common Voice platform[14]. Regular data pulls from the Awal platform ensure the repository stays current with new contributions. Real-time data metrics are available on the Awal homepage.

### 3.6. *Limitations, challenges and way forward*

After eighteen months of community engagement, the Awal project has revealed several challenges that impact participation and data collection effectiveness. Through interviews with promoters and contributors, we identified key barriers that explain why the initiative has not reached the

12 Common Voice Corpus 22.0 release

13 https://huggingface.co/datasets/collectivat/amazic

14 https://commonvoice.mozilla.org/en/datasets

participation levels initially expected. Our analysis revealed six main conclusions:

1. **General tamazight-speaking public don't trust their reading and writing abilities.** Most people don't know Tifinagh and never had the chance to learn it. Despite being fluent speakers, many community members expressed deep uncertainty about their writing abilities. Even those who regularly write Tamazight in Latin form in family messaging groups or social media view their informal writing as inadequate for data collection purposes. Participants in workshops often said they "don't write correctly" or "don't know the standard form," leading them to exclude themselves from contributing to a space where their writing will be recorded and seen by others.
2. **Micro-translation tasks emerge as a promising approach for data creation.** Current activist-linguist efforts focus on encouraging people to write, and translation emerges as a promising approach for content creation. Rather than asking participants to generate original text, providing source material in other languages—or even AI-generated content—removes the creative barrier that prevents many from contributing. This approach has shown success in platforms like Tatoeba, where collaborative translation has resulted in substantial Tamazight content across multiple varieties. The Awal platform's automatic sentence loading feature addresses this obstacle by eliminating the "what should I write?" problem that blocks participation.
3. **Most contributions come from academic, linguist and activist community members.** Even though we received many positive responses from the general public during events, most impact in terms of data volume came from activists and academics who already understood the importance of language digitization—people already working within cultural associations, researchers, or those with formal linguistic training. The crowdsourcing model assumes a level of literacy and standardization that does not match the current reality of the general Tamazight-speaking community, particularly in the diaspora. The project should focus on this specialized public if it wants to generate impact in terms of data size, while maintaining outreach to the general public so that more people become knowledgeable about technology and motivated to develop their reading and writing skills.
4. **There's no consensus on representing dialects on the platform.** The project aimed to welcome dialectal diversity while maintaining

data quality through a contribution labeling system allowing users to mark their dialect. But this approach revealed tensions within the Tamazight linguistic ecosystem. Standard Moroccan Tamazight remains unfamiliar to many diaspora speakers and those who learned the language orally at home. Meanwhile, the academic community recommends not using any dialect labels at all, considering there's already unity in written form.

5. **Code-mixing and linguistic purity present challenges for data quality.** Diaspora communities often incorporate terms from their residence languages, while speakers in Morocco frequently mix Darija into their Tamazight. Finding truly Amazigh equivalents requires extensive research across dialects and regions, including consulting variants from other countries—a task that demands linguistic expertise rather than casual knowledge. We observed that elderly women in the diaspora maintained purer dialectal forms compared to younger speakers, suggesting that language contact effects intensify across generations and geographic distance from origin communities.
6. **A more internationalist approach could generate more impact.** The promotional materials primarily in Catalan limited reach to the broader international Amazigh diaspora in Belgium, France, and other regions, and also prevented reaching the population in Morocco, which would help scale participation massively. This would require collaboration with entities beyond Catalonia, which the current scope doesn't allow.

Despite these challenges, the project has demonstrated valuable pathways forward. The positive reception and widespread awareness generated by Awal indicates strong community interest in language technology development. Many people, thanks to this initiative, were surprised and motivated by the idea that advanced technologies can also exist in their native language.

As a way forward, the most promising approach involves concentrating efforts on academic and activist communities who possess both linguistic expertise and technological literacy for digital language preservation. These contributors can build foundational datasets while the project develops complementary pedagogical components that build writing confidence in the broader community. Additionally, adapting the platform to accept audio input alongside text could reduce the complexities of having contributions on two different platforms. The challenges identified here also point toward the need for longer-term institutional partnerships with educational institutions in Morocco, where standardization efforts are more advanced and writing instruction is integrated into formal education[15].

## 4. Conclusion

The Awal project demonstrates that community-driven language technology development for underrepresented languages requires adaptation to local linguistic realities and cannot simply replicate models designed for standardized languages. While our 18-month experiment collected meaningful data and raised community awareness, it revealed fundamental tensions between crowdsourcing assumptions and the complex sociolinguistic context of Tamazight speakers. In parallel work, we are curating and correcting standard MT datasets and fine-tuning MT models using the collected data. As a public-funded diaspora initiative, Awal highlights the need for cross-border collaboration that reflects the transnational reality of Tamazight-speaking communities. We call for greater institutional support that combines grassroots community efforts with formal language planning, moving beyond isolated initiatives toward coordinated language technology development that serves speakers across borders and generations.

## 5. Acknowledgements

This article was written as a result of research conducted under CIEMEN's and Fundació pels Drets Col·lectius dels Pobles' Som Part project, which received funding from the Catalan Agency for Development Cooperation (ACCD) and the Municipality of Barcelona.

We extend our sincere gratitude to Lalla Ghizlan Baryala, Brahim Essaidi, Mohamed Aymane Farhi, and Naceur Jabouja for their invaluable assistance and insights that contributed to the development of this paper.

We also thank all the volunteers who contributed to the Awal platform and participated in our workshops, sharing their experiences and insights about their knowledge in Tamazight and technology.

15 Despite multiple attempts to establish collaboration with IRCAM, we received no response from the institution.